\documentclass[letterpaper, 10 pt, conference]{ieeeconf}  

\IEEEoverridecommandlockouts                              

\usepackage[utf8]{inputenc}
\usepackage{todonotes}
\usepackage{amsfonts}
\usepackage{tabularx}
\usepackage{array, booktabs, makecell}
\usepackage{siunitx}
\usepackage{tablefootnote}
\usepackage{lipsum}

\usepackage{times}

\usepackage{multicol}

\usepackage{pdfsync}
\usepackage{graphicx}
  \graphicspath{{figure/}}
  \DeclareGraphicsExtensions{.pdf,.jpg,.png,.eps}
\usepackage{pdfpages}
\usepackage{bm}
\usepackage{amsmath,amssymb}
\usepackage{algorithm}
\usepackage{algorithmicx}
\usepackage{algpseudocode}
\usepackage{xspace}
\usepackage{booktabs}
\usepackage{soul}
\usepackage{multirow}
\usepackage{todonotes}

\renewcommand{\eqref}[1]{(\ref{#1})}
\newcommand{\figref}[1]{Fig.~\ref{#1}}

\newcommand{\subfig}[1]{\textit{#1}}
\newcommand{\tabref}[1]{Table~\ref{#1}}

\newcommand{\ie}{\textrm{i.e.}}
\newcommand{\eg}{\textrm{e.g.}}
\newcommand{\etc}{\textrm{etc.}}

\DeclareMathOperator*{\argmin}{arg\,min} 

\newlength{\citeskipup}
\newlength{\citeskipdown}

\usepackage{color}
\definecolor{fullred}{rgb}{0.95,.0,.1} 
\newcounter{cmt}

\newcommand{\simname}{SUMMIT\xspace}
\newcommand{\algname}{Context-POMDP\xspace}
\newcommand{\modelname}{Context-GAMMA\xspace}

\DeclareSIUnit\Ms{m/s}

\newcommand{\norm}[1]{\left\lVert#1\right\rVert}
\usepackage{hyperref}
\usepackage[font=small,labelfont=bf,tableposition=top]{caption}

\makeatletter
\newcommand{\printfnsymbol}[1]{%
  \textsuperscript{\@fnsymbol{#1}}%
}
\makeatother

\newcommand\blfootnote[1]{%
  \begingroup
  \renewcommand\thefootnote{}\footnote{#1}%
  \addtocounter{footnote}{-1}%
  \endgroup
}

\mathchardef\mhyphen="2D
\title{\simname: A Simulator for Urban Driving \\in Massive Mixed Traffic}
\author{Panpan Cai\printfnsymbol{1}, Yiyuan Lee\printfnsymbol{1}, Yuanfu Luo, David Hsu
}

\begin{document}
   
\twocolumn[{%
\renewcommand\twocolumn[1][]{#1}%
\maketitle


\begin{center}
\setlength{\fboxrule}{0pt}
\centering                                                              \begin{tabular}{ccc}
\fbox{\includegraphics[width=0.3\textwidth]{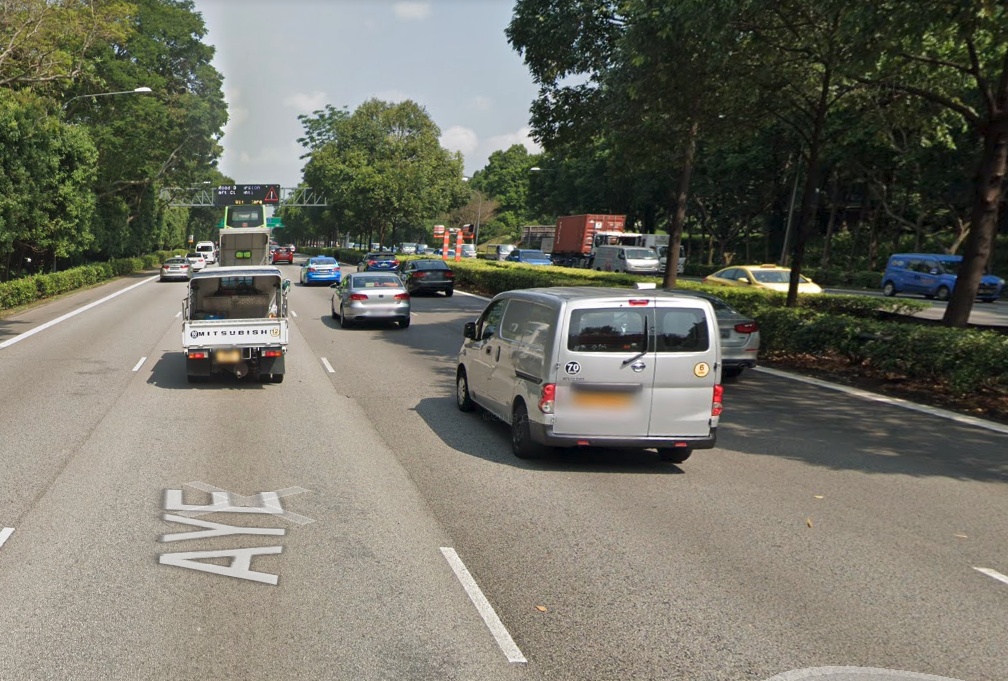}} & \fbox{\includegraphics[width=0.3\textwidth]{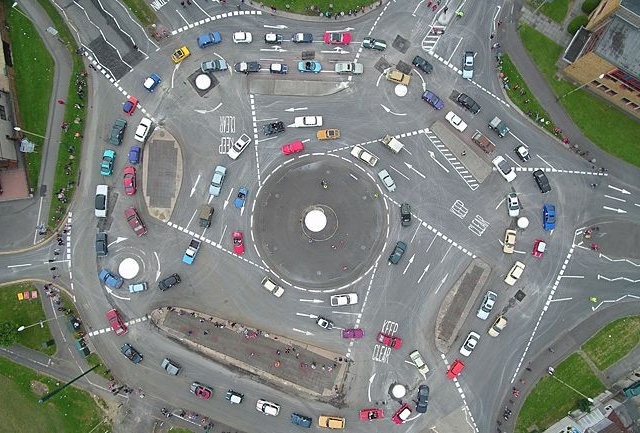}} & 
\fbox{\includegraphics[width=0.3\textwidth]{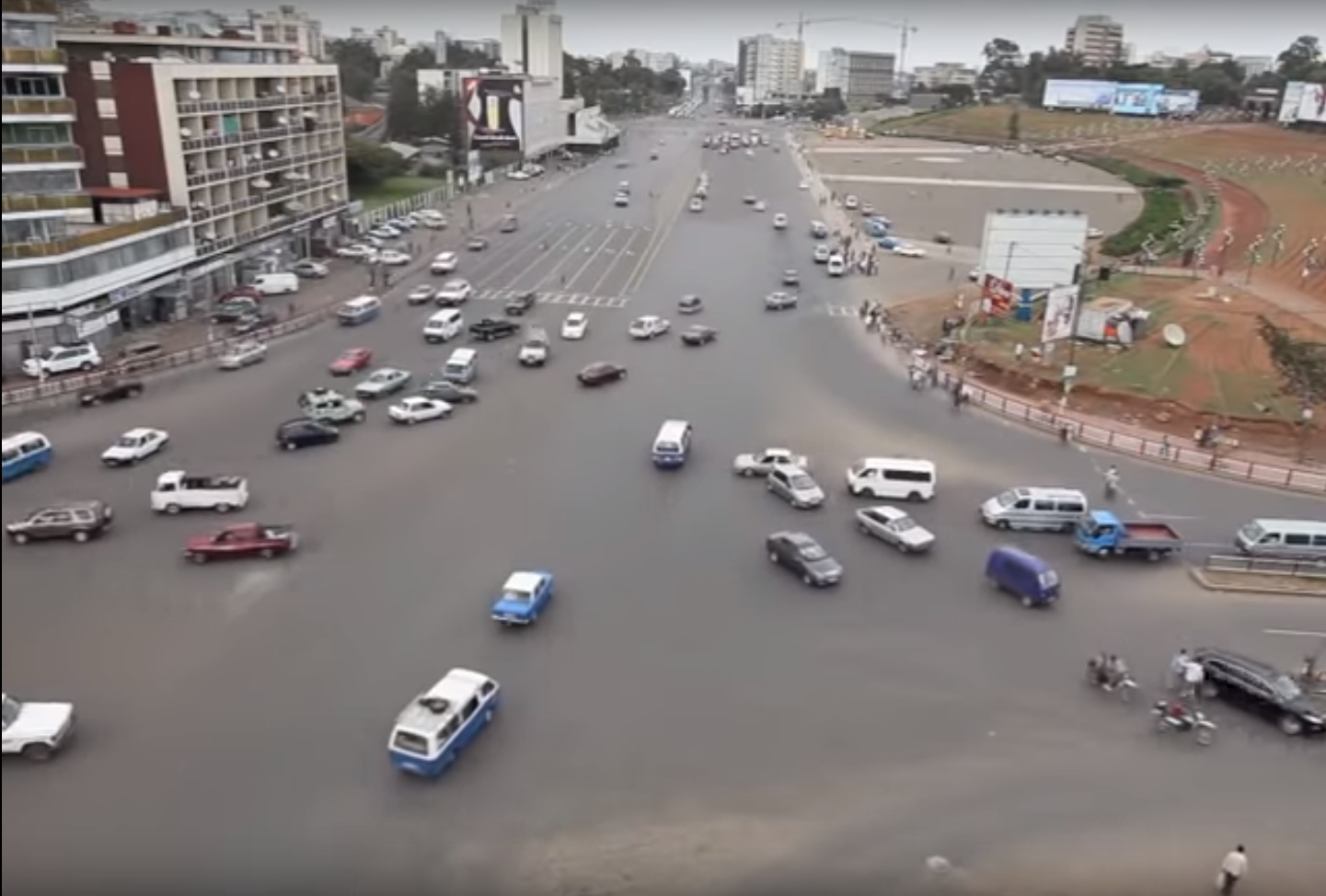}} \\
(\subfig{a}) Singapore-Highway & (\subfig{b}) Magic-Roundabout & (\subfig{c}) Meskel-Intersection\\    
\fbox{\includegraphics[width=0.3\textwidth]{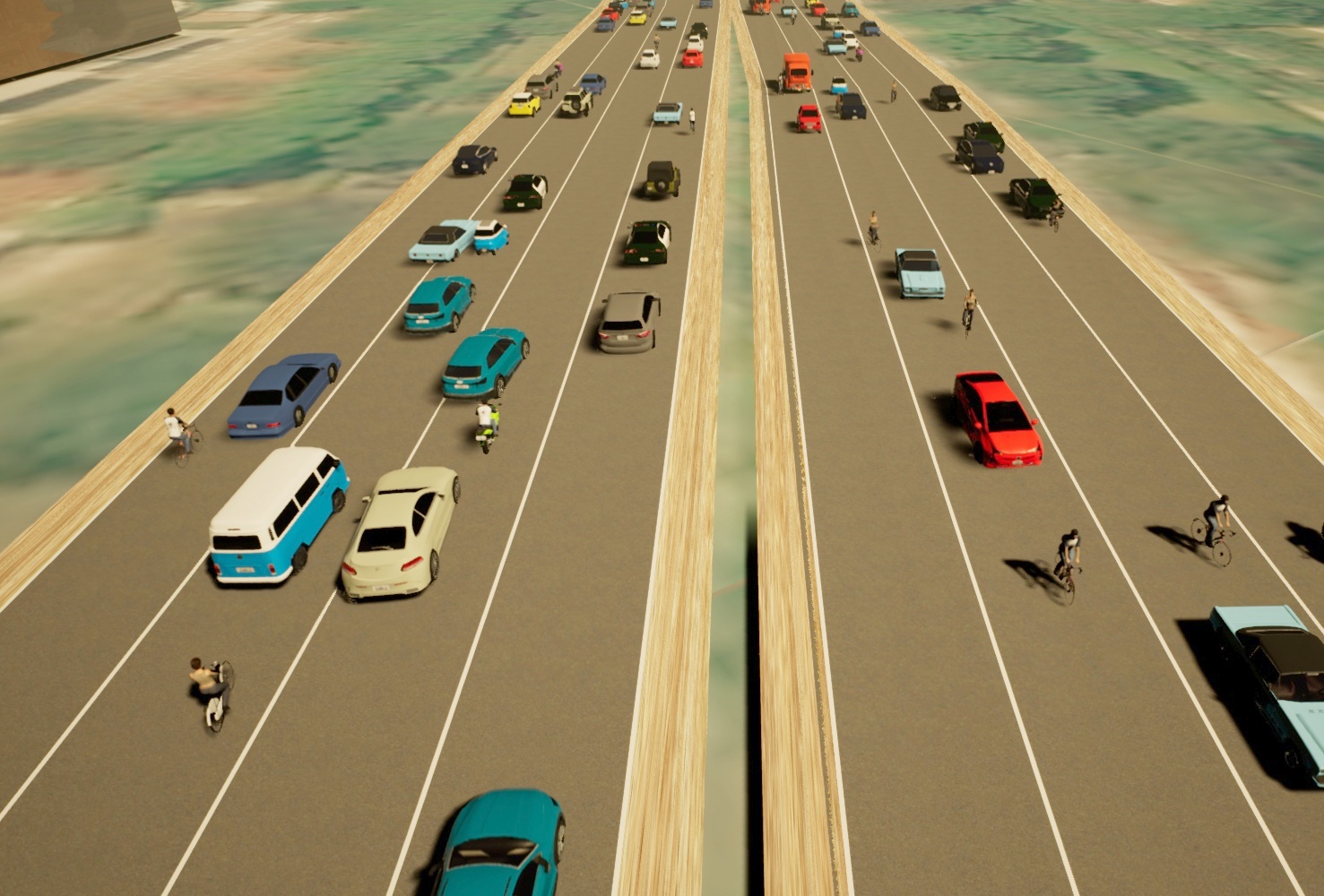}} &  \fbox{\includegraphics[width=0.3\textwidth]{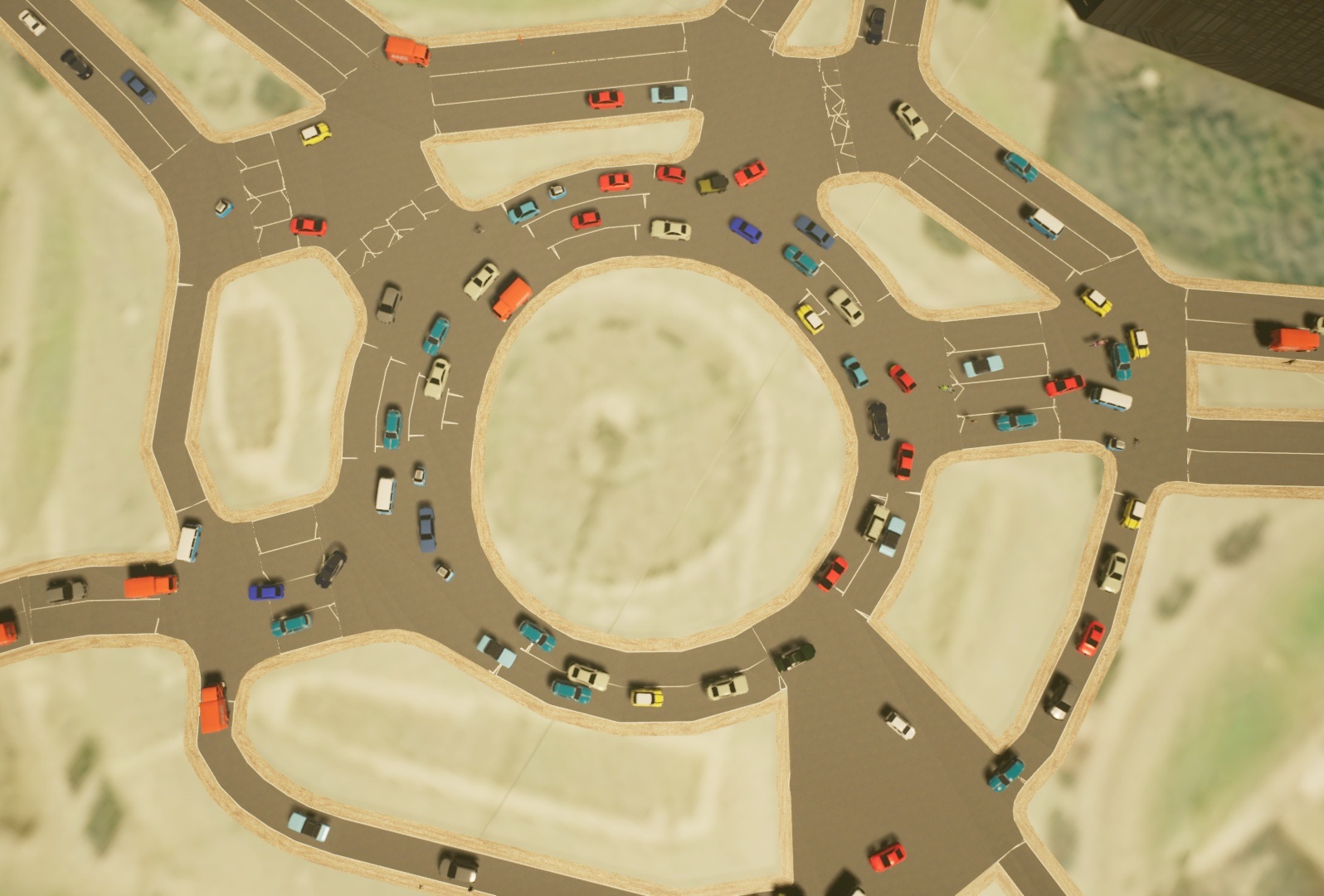}} &  \fbox{\includegraphics[width=0.3\textwidth]{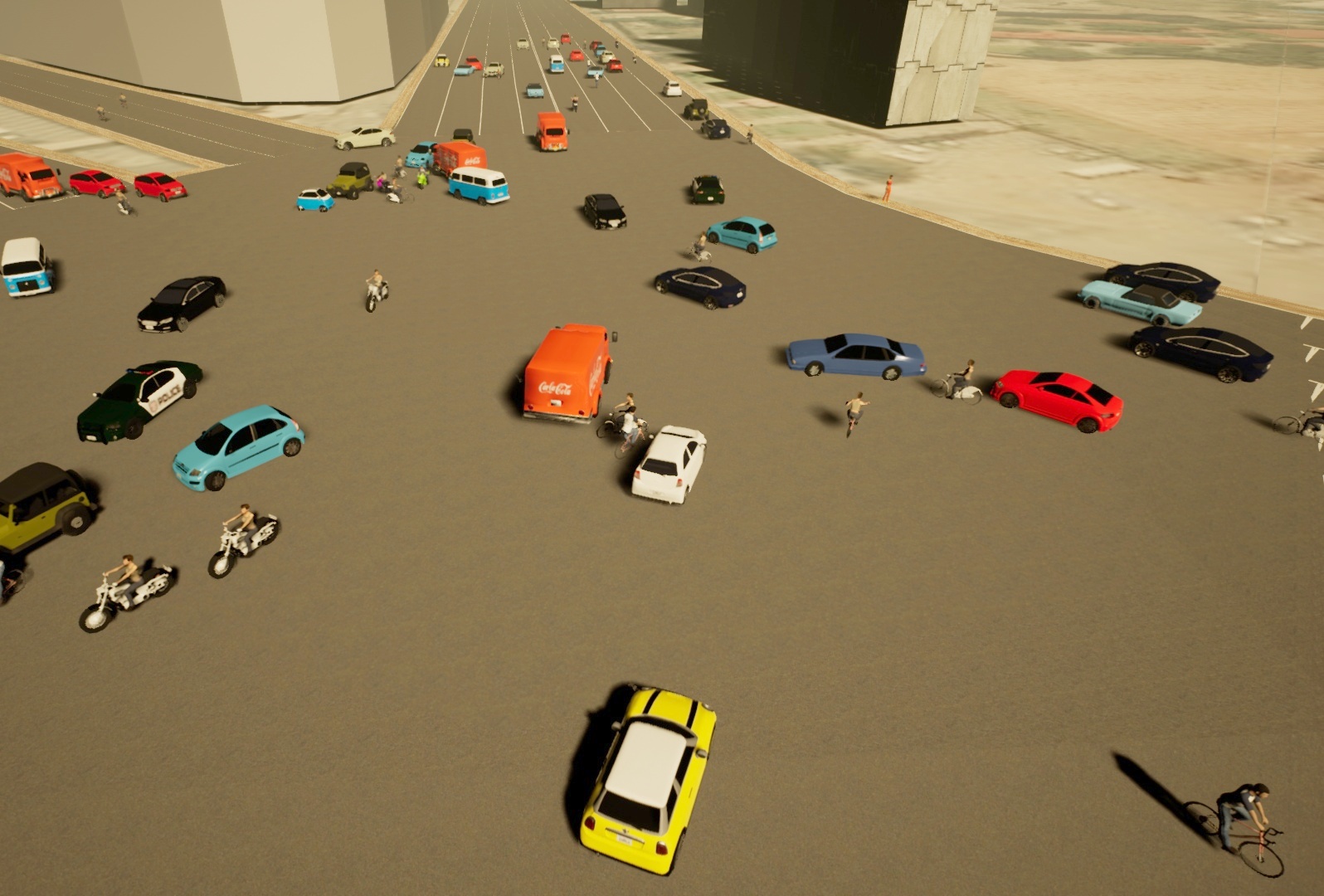}}  \\
(\subfig{d}) Singapore-Highway in \simname & (\subfig{e}) Magic-Roundabout in \simname & (\subfig{f}) Meskel-Intersection in \simname\\    
\end{tabular}  
\captionof{figure}{ Benchmark scenes in the real world and corresponding scenes in SUMMIT.
}        
\label{fig:benchmarks}     
\end{center} 
}]

\begin{abstract}
Autonomous driving in an unregulated urban crowd is an outstanding challenge,
especially, in the presence of many aggressive, high-speed traffic
participants.  This paper presents \simname, a high-fidelity simulator that
facilitates the development and testing of crowd-driving algorithms.  By
leveraging the open-source OpenStreetMap map database and a heterogeneous
multi-agent motion prediction model developed in our earlier work, \simname
simulates dense, unregulated urban traffic for heterogeneous agents at any
worldwide locations that OpenStreetMap supports. \simname is built as an
extension of CARLA and inherits from it the physics and visual realism for
autonomous driving simulation.  \simname supports a wide range of
applications, including perception, vehicle control and planning, and end-to-end
learning. We provide a context-aware planner together with benchmark scenarios
and show that \simname generates complex, realistic traffic behaviors in 
challenging crowd-driving settings. Code for the simulator and the planner are available at \href{https://github.com/AdaCompNUS/summit}{https://github.com/AdaCompNUS/summit} and \href{https://github.com/AdaCompNUS/Context-POMDP}{https://github.com/AdaCompNUS/Context-POMDP}.
\end{abstract}

\section{Introduction}

The vision of using autonomous driving to improve the safety and convenience of our daily life is coming closer. However, driving in \textit{unregulated}, crowded urban environments, like in uncontrolled roads or unsignalised intersections in less-developed countries (\figref{fig:benchmarks}), remains an open problem. Human participants can be fairly aggressive in these scenarios. One may disregard or be unaware of traffic rules, leading to behaviors like close following, inappropriate overtaking, illegal turning and crossing, etc. The road condition can become highly chaotic when involving many participants.
Technical challenges for driving in unregulated urban crowds come from the complexity of both crowd behaviors and map environments. Traffic agents can be significantly different from each other. Cars, buses, bicycles, and motorcycles have different geometry, kinematics, and dynamics. Human participants also have different behavioral types - being conservative or aggressive, attentive or distracted, etc.
In terms of the map environment, urban roads can have complex and versatile layouts: multi-lane roads, intersections, roundabouts, etc. Road structures significantly influence the motion of traffic agents and thus generate very different crowd behaviors in different locations. Such environments raise enormous difficulties for perception, control, planning, and decision-making of robot vehicles. \blfootnote{\printfnsymbol{1}The authors contributed equally.}
\blfootnote{The authors are with School of Computing, National University of Singapore, 117417 Singapore. \tt\footnotesize{\{caipp, leeyiyuan, yuanfu, dyhsu\}@comp.nus.edu.sg}.}

High-quality data for developing, training, and testing crowd-driving algorithms are, however, difficult and expensive to acquire due to the cost of devices, regulations and safety constraints. Although there are publicly available data sets like  KITTI \cite{KITTI}, BDD100K \cite{BDD100K}, Oxford RobotCar \cite{Oxford}, \etc, that provide real-world driving data with rich sensor inputs, these data are not \emph{interactive}, \ie, one cannot model the reactions of exo-agents to the robot's decisions. Such data are, however, extremely important for robust planning and learning. 
A promising source of interactive data are driving simulators that can offer a virtually unlimited amount of controllable scenarios. However, existing driving simulators do not capture the full complexity of unregulated urban crowds such as complex road structures and traffic behaviors, and are thus insufficient for testing or training robust driving algorithms. We aim to fill this gap.

We develop a new simulator, \simname \footnote{Code available at \href{https://github.com/AdaCompNUS/summit}{https://github.com/AdaCompNUS/summit}}, that generates high-fidelity, interactive data for unregulated, dense urban traffic on complex real-world maps. 
\simname uses real-world maps fetched from online sources to provide a virtually unlimited source of complex environments. Given arbitrary locations, the simulator automatically generates crowds of heterogeneous traffic agents with sophisticated, unregulated behaviors. The simulator leverages road contexts of real-world maps to guide the behaviours of traffic agents topologically and geometrically in order to construct realistic traffic conditions. We implemented \simname based on CARLA \cite{CARLA} to leverage the high-fidelity physics, rendering, and sensors. Through a python-based API, \simname reveals rich sensor data, semantic information, and road contexts to external algorithms, enabling the application in a wide range of fields such as perception, vehicle control and planning, end-to-end learning, etc. 
We provide both qualitative and quantitative results to show that \simname can generate complex, realistic mixed traffic in real-world urban environments. 

We further provide a context-aware planner, \algname \footnote{Code available at \href{https://github.com/AdaCompNUS/Context-POMDP}{https://github.com/AdaCompNUS/Context-POMDP}}, as a reference for future crowd-driving algorithms. The planner explicitly reasons about interactions among traffic agents under the uncertainty of human intentions and driving types. By further conditioning planning on available road contexts, \algname achieves safe and efficient driving in very challenging scenarios. 


\section{Related Work}
\begin{table}[!t]
\centering
\caption{Comparison between \simname and existing driving simulators.}
\begin{tabular}{ccccc }
\hline
\thead{Simulator} & \thead{Real\\-world \\
Maps} & \thead{Unregulated\\behaviors} & \thead{Dense\\Traffic} \tablefootnote{We only check-mark simulators explicitly featuring crowd behaviours.} &  \thead{Realistic\\Visuals \&\\Sensors} \\
\toprule
SimMobilityST \cite{SimMobilityST} & $\checkmark$ & $\times$ & $\checkmark$  & $\times$   \\
SUMO \cite{SUMO}&  $\checkmark$  &$\times$ & $\checkmark$ & $\times$ \\
TORCS \cite{TORCS}  & $\times$ & $\checkmark$ & $\times$ & $\checkmark$ \\
Apollo \cite{Apollo} & $\times$  & $\times$ & $\times$  & $\checkmark$\\
Sim4CV \cite{Sim4CV} & $\times$ & $\times$ & $\times$ & $\checkmark$ \\
GTAV \cite{GTAV}  & $\times$ & $\times$ & $\times$  & $\checkmark$ \\
CARLA \cite{CARLA}  & $\times$  & $\times$ & $\times$ & $\checkmark$ \\
AutonoViSim \cite{AutonoViSim} & $\times$ & $\checkmark$ & $\checkmark$ & $\checkmark$  \\
Force-based \cite{chao2019force} & $\times$ & $\times$ & $\checkmark$ & $\checkmark$  \\
\simname (ours) & $\checkmark$  & $\checkmark$ & $\checkmark$ & $\checkmark$\\
\bottomrule

\end{tabular}
\label{tab::simulators}
\vspace{-0.5cm}
\end{table}

\subsection{Existing Driving Simulators}
Driving simulators have been extensively applied to boost the development of autonomous driving systems.
Recent simulators (\tabref{tab::simulators}) have brought realistic visuals and sensors, but do not capture the complexities of urban environments and unregulated traffic behaviors.

Multi-car simulators like TORCS \cite{TORCS,Fluids,CoInCar-Sim} focus on interactions between multiple robot-vehicles. These simulators suit the study of complex interactions between agents, but can hardly scale up to crowded urban scenes.
CARLA \cite{CARLA}, Sim4CV \cite{Sim4CV}, and GTA \cite{GTAV} explicitly feature detailed physics modeling and realistic rendering for end-to-end learning. CARLA also provides a rich set of sensors such as cameras, Lidar, depth cameras, semantic segmentation, etc. However, these simulators rely on predefined maps, limiting the variety of environments. The simulated traffic also have relatively low density and simple rule-based behaviors. 
Another class of simulators \cite{SUMO,SimMobilityST,AutonoViSim,chao2019force} feature traffic simulation and control in urban environments. Among them, SUMO \cite{SUMO} and SimMobilityST \cite{SimMobilityST} support real-world maps but use simple rule-based behaviors, while another class \cite{AutonoViSim,chao2019force} apply more sophisticated motion models but are restricted to predefined maps. We aim to model the complexities in both urban maps and traffic behaviors in an automatic and unified framework.

\subsection{Crowd Simulation Algorithms} 
Existing crowd simulation algorithms, \eg, social force and velocity obstacles, can in principle be applied to generate crowd behaviors in urban environments. 
Social force \cite{helbing1995social,lohner2010modeling,ferrer2013robot,pellegrini2009you} assume that traffic-agents are driven by attractive forces exerted by the destination and repulsive forces exerted by obstacles. Social force can simulate large crowds, but the quality of interactions are constrained by model simplicity. 
Velocity Obstacle (VO) \cite{fiorini1998motion} and Reciprocal Velocity Obstacle (RVO) \cite{van2008reciprocal,van2011reciprocal,snape2011hybrid} compute collision free motion by optimizing in the feasible velocity space. Variants such as GVO \cite{wilkie2009generalized}, NH-ORCA \cite{alonso2013optimal}, B-ORCA \cite{alonso2012reciprocal}, PORCA \cite{PORCA} explicitly handle non-holonomic traffic agents. Some variants model behavioral types of crowd agents such as patience \cite{PORCA} and attention \cite{cheung2018efficient}. A recent model, GAMMA \cite{GAMMA}, can simulate heterogeneous traffic agents with different geometry, kinematics, and behavioral types in a unified velocity-space framework. 
The behavior model in \simname extends the framework of GAMMA to encode topological road contexts such as lanes and pedestrian sidewalks to closely represent real-world scenarios.

\section{\simname Simulator}
\begin{figure*}[!t]
\centering
\includegraphics[width=0.9\textwidth]{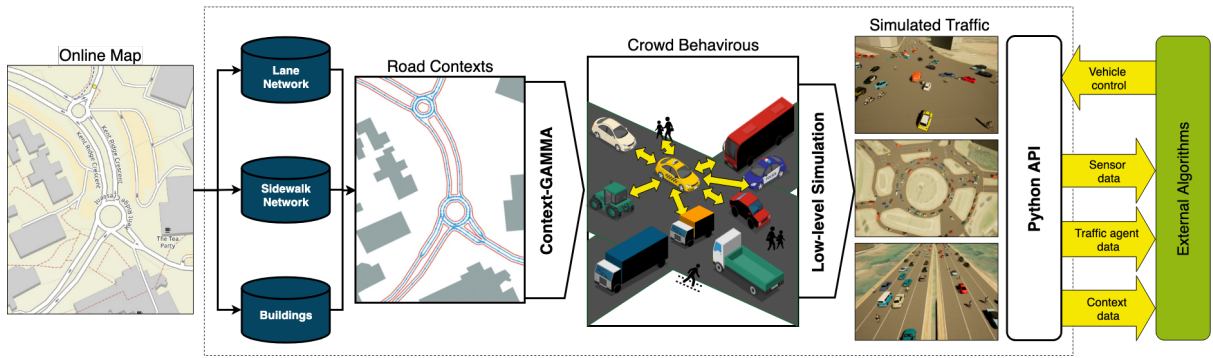}
\caption{An overview of \simname that simulates massive mixed traffic at any location in the world.}
\label{fig::overview}

\end{figure*}
\simname focuses on simulating complex unregulated behaviours of dense urban traffic in complex real-world maps. It is designed for generating high-fidelity interactive data to facilitate the development, training, and testing of crowd-driving algorithms.
\simname automatically generates massive mixed traffic using topological road contexts and optimization-based unregulated crowd behaviors. 
\simname fetches real-world maps from the OpenStreetMap \cite{OSM}, and constructs two topological graphs: a lane network for vehicles, and a sidewalk network for pedestrians. These networks form a representation of the road contexts. 
Then, our behavior model, \modelname, takes road contexts as input to guide the traffic behaviors geometrically and topologically. At the microscopic level, \modelname uses velocity-space optimization to generate sophisticated, unregulated crowd behaviors.
The low-level structure of \simname is based on CARLA, retaining its desirable features such as high-fidelity physics, realistic rendering, weather control, and rich sensors. \figref{fig::overview} provides an overview of \simname.

\subsection{Representing Real-world Maps}
\subsubsection{Lane Network}
A lane network in \simname defines the connectivity of the road structure at the fidelity of individual lanes. The network consists of directed lane segments and connections between them. \simname relies on SUMO \cite{SUMO} to automatically convert OSM maps to lane networks. The extensive suite of network editing tools provided by SUMO can also be leveraged to improve and customize maps. The lane network interface allows users to locate traffic agents on the lane network and retrieve connected lane segments. The interface closely follows CARLA's waypoint interface, so that CARLA users can easily adapt to it.



\subsubsection{Sidewalk Network}
A sidewalk network in \simname defines the behaviors of pedestrians, which usually walk along road edges and occasionally cross roads. The network contains sidewalks near road edges defined as poly-lines and connections between sidewalks defined as cross-able roads. The sidewalk poly-lines are extracted from the geometry of roads. Similar to the lane network, the sidewalk network interface allows users to locate pedestrians on the network and retrieve the opposite sidewalk for road-crossing.


\subsubsection{Occupancy Map Interface}
We additionally provide an occupancy map interface to expose drive-able regions for the ego-vehicle. An occupancy map is the top-down projection of the road geometry, aligned with the ego-vehicle's location and heading direction. It can be used either for collision checking in control and planning algorithms or as bird-view input to neural networks.



\subsubsection{Landmarks}
\simname also makes use of landmark data in OSM maps such as buildings and forests to provide structurally rich and realistic visuals. We additionally support randomization of the landmark textures to generate more versatile visual inputs and enable techniques such as domain randomization \cite{domain}.



\subsection{Crowd behavior Modelling}
\simname uses \modelname, a context-aware crowd behavior model, to generate sophisticated interactive behaviors of traffic agents. \modelname extends GAMMA \cite{GAMMA} to incorporate road contexts and models them as constraints in velocity space. The realism and accuracy of GAMMA has been validated in various real-world datasets. For completeness, we briefly introduce GAMMA, and present the extensions in \modelname.

\begin{figure}
\centering
\begin{tabular}{cc}
\hspace{-7pt}
\includegraphics[height=0.12\textwidth]{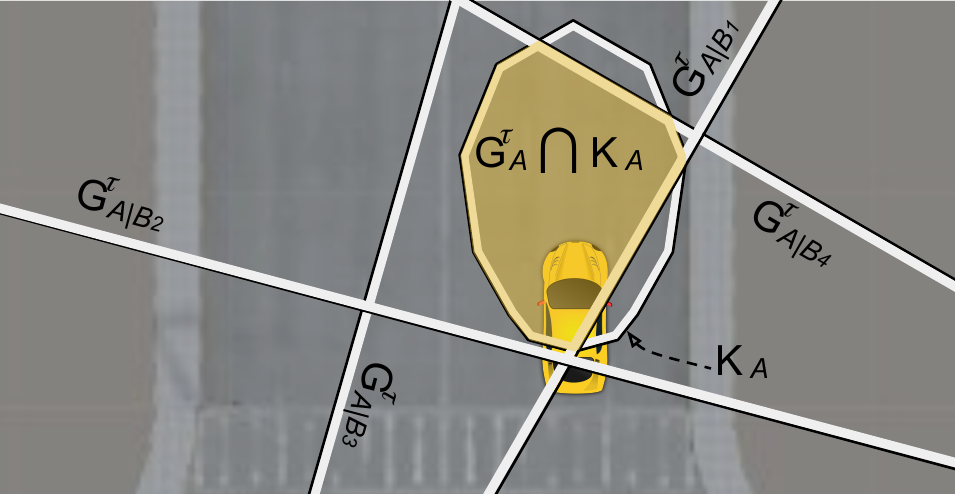}&
\hspace{-10pt}
\includegraphics[height=0.12\textwidth]{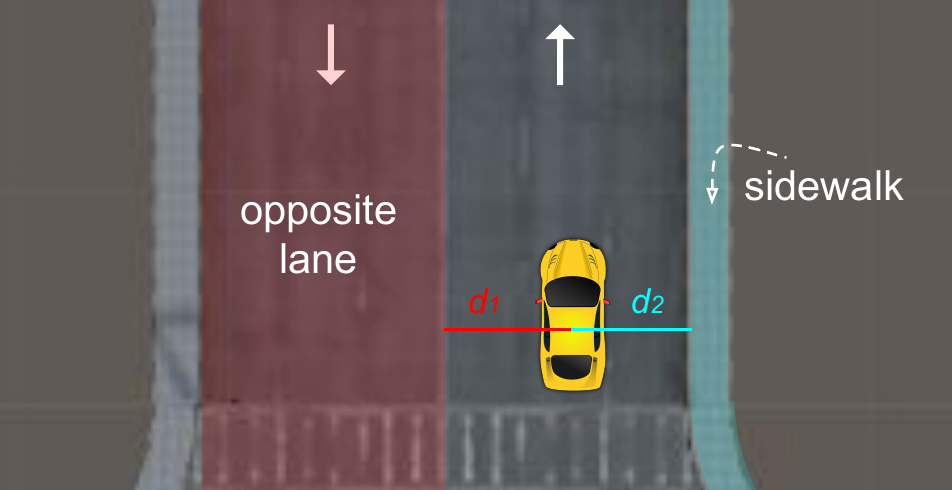} \\  
\hspace{-7pt} (\subfig a) &  \hspace{-10pt}(\subfig b) \\
\hspace{-7pt}
\includegraphics[height=0.12\textwidth]{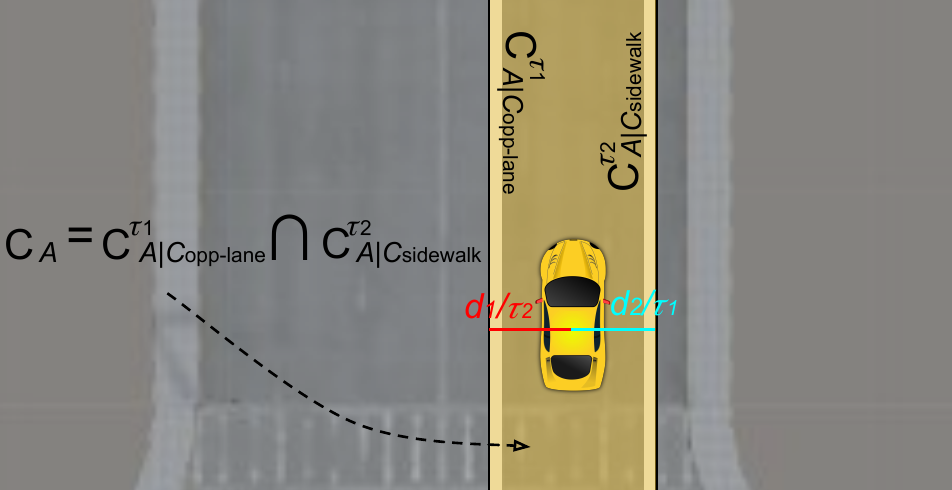}&
\hspace{-10pt}
\includegraphics[height=0.12\textwidth]{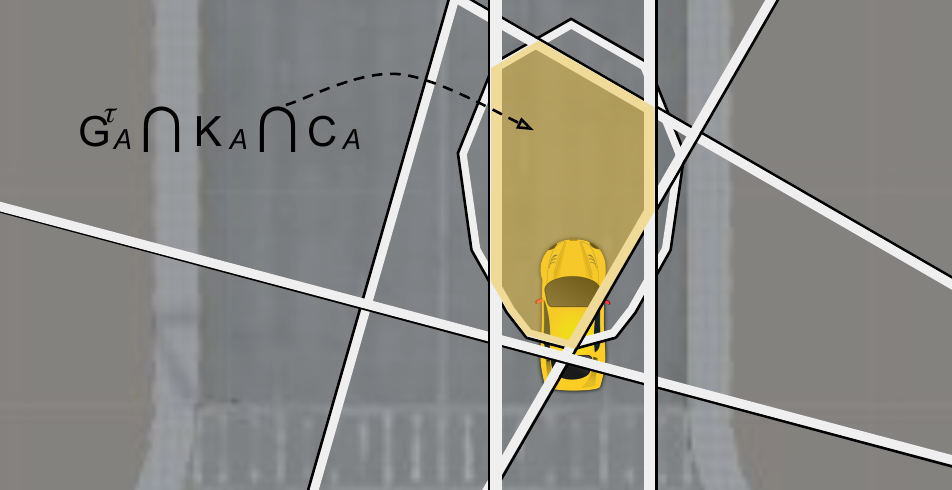} \\
\hspace{-7pt} (\subfig c) &  \hspace{-10pt}(\subfig d) \\
\end{tabular}   
\caption{Crowd behaviour modelling with \modelname: (\subfig a) the feasible velocity space (yellow) of an agent $A$ as the intersection of kinematic constraints $\mathrm{K}_A$ and geometry constraints $\mathrm{G}_A^\tau$. (\subfig b) A car constrained by the opposite lane and the sidewalk. (\subfig c) The corresponding contextual constraints $\mathrm{C}_{A|C_{\mathrm{opp\mhyphen lane}}}^{\tau_1}$ and $\mathrm{C}_{A|C_{\mathrm{sidewalk}}}^{\tau_2}$. (\subfig d) The augmented feasible velocity space.}
\label{fig:gamma}
\vspace{-0.6cm}
\end{figure}

GAMMA formulates the motion of traffic agents as constrained geometric optimization in velocity space. It assumes that each traffic agent optimizes its velocity based on the navigation goal, while being constrained by kinematic constraints (e.g. non-holonomic motion of car) and geometric constraints (collision avoidance with nearby agents). For a given agent $A$, let $\mathrm{K}_A$ represent the set of velocities that satisfy kinematic constraints and $\mathrm{G}_A^\tau$ represent the set of velocities that satisfy geometric constraints for at least $\tau$ time. Then GAMMA selects for $A$ a new velocity from their intersection:
\begin{equation}\label{eq:gamma_obj_fun}
v_A^\mathrm{new} = \argmin_{v \in \mathrm{G}_{A}^{\tau} \cap \mathrm{K}_{A}} \norm{v - v_A^\mathrm{pref}},
\end{equation}
where $v_A^\mathrm{pref}$ is $A$'s preferred velocity computed from its goal. When computing $\mathrm{K}_A$ and $\mathrm{G}_A^\tau$, GAMMA also takes into account responsibility and attention of the agent to generate more human-like motions. We refer readers to \cite{GAMMA} for more details of the construction of $\mathrm{K}_A$ and $\mathrm{G}_A^\tau$. Geometrically, $\mathrm{K}_A$ is a convex velocity set and $\mathrm{G}_A^\tau$ is the intersection of velocity-space half planes. See \figref{fig:gamma}\subfig{a} for an example of $\mathrm{K}_A$ and $\mathrm{G}_A^\tau$.

GAMMA handles heterogeneous traffic agents with different kinematics and geometry in a unified velocity-space framework. It has been proven experimentally to accurately predict behaviors of real-world traffic agents \cite{GAMMA}. 

However, as GAMMA does not make explicit use of road contexts, it often fails to generate realistic simulation for complex urban roads. GAMMA agents can aggressively head towards their goals and be trapped by the complex road structure. In the real-world, road contexts can effectively guide and constrain traffic agents' behaviors: vehicles tend to follow particular lanes when the road is clear and avoid to drive along the wrong direction. Moreover, as traffic agents are heterogeneous, they are affected by different sets of static obstacles: pedestrians consider sidewalks as open spaces, but vehicles consider them as obstacles.

Our new model, \modelname, provides a general way to embed road contexts as objectives and constraints in velocity space.
\modelname extracts the preferred velocity $v^{\mathrm{pref}}$ of traffic agents from the lane and sidewalk networks. To diversify agent behaviors, it randomly selects a lane from all feasible lanes ahead of the agent, and points $v^{\mathrm{pref}}$ to a look-ahead waypoint along the selected lane.

\modelname can also model traffic rules by casting contextual constraints, \eg, no wrong-direction driving, into half planes in agents' velocity space, forcing them to select velocities complying with the road rule. Denote the intersection of all the contextual half planes for an agent $A$ as $\mathrm{C}_A$. \modelname optimizes the agent velocity in the augmented feasible velocity space $\mathrm{K}_A \cap \mathrm{G}_A^\tau \cap \mathrm{C}_A$. Note that $\mathrm{K}_A \cap \mathrm{G}_A^\tau \cap \mathrm{C}_A$ is also convex by construction and the objective function \eqref{eq:gamma_obj_fun} is quadratic. Therefore, the optimization problem can be efficiently solved in linear time.

\figref{fig:gamma}\subfig{b} and \figref{fig:gamma}\subfig{c} show an example of contextual constraints. To prevent car $A$ in \figref{fig:gamma}\subfig{b} from driving to the opposite lane within a time window $\tau_1$, the lateral speed of the car should be constrained under $d_1/\tau_1$, where $d_1$ is the distance from the car to the opposite lane. This constraint forms a half-plane in the velocity space, $\mathrm{C}_{A|C_{\mathrm{opp \mhyphen lane}}}^{\tau_1}$, defined by a separation line parallel to the opposite lane with an offset of $d_1/\tau_1$ from the origin. Any velocity in $\mathrm{C}_{A|C_{\mathrm{opp \mhyphen lane}}}^{\tau_1}$ would be feasible. 
Similarly, collision avoidance with the sidewalk can result in another half-plane $\mathrm{C}_{A|C_{\mathrm{sidewalk}}}^{\tau_2}$ in the velocity space with an offset of $d_2$ from the origin.
The intersection of the two half-planes forms $A$'s contextual constraint, $\mathrm{C}_{A}$ (\figref{fig:gamma}\subfig{c}), which is further imposed on $\mathrm{K}_A \cap \mathrm{G}_A$ to form the feasible space of $A$ (\figref{fig:gamma}\subfig{d}).
\subsection{Interfaces}
The Python API of \simname extends that of CARLA, exposing to external algorithms not only sensor data and agent states, but also road contexts like lane networks, sidewalk networks, and map occupancy grids. Algorithms can also send vehicle control back to the simulation including steering, acceleration, braking, reversing, \etc. \simname thus enables a wide range of applications such as perception, sensor-based control, model-based reasoning, and end-to-end learning.

\section{Context-aware POMDP Planning}
\simname also offers an expert planner for autonomous driving. 
Planning for driving in an unregulated dense traffic is extremely challenging. The robot has to be smart enough to make efficient progress, instead of being "frozen" and stuck in the crowd. In the meantime, a highly dynamic and interactive crowd makes the task safety critical. Mistakes in planning can lead to severe or even fatal accidents.

The key to success is to explicitly model interactions among agents as well as the uncertainty on human inner states. Such sophisticated planning requires long-term reasoning in the belief-space, which brings combinatorial complexities.
To make the problem tractable, we propose to condition planning on road contexts.
We formulate driving as a context-aware POMDP, and solves it efficiently using online belief tree search \cite{Hyp-despot}. We refer to this planner as \algname.

\algname conditions Monte Carlo simulations for the future on both human hidden states and road contexts. This is achieved by forward simulating exo-agents using \modelname.
\algname consists of two components: a belief tracker that infers a joint belief over exo-agents' hidden states, and an online POMDP solver that computes an optimal driving action for the current belief.
\subsection{Belief Tracking}
The belief tracker maintains a joint belief over two dimensions of hidden states: 
\begin{itemize}
  \item The \textit{intention} of the traffic agent: Let $U_i,i\in I_{exo}$ be the set of path candidates for the $i$th traffic agent extracted from the road contexts such as the lane network and the sidewalk network. This agent may take any of the path candidates in $U_i$ as its actual intention. 
  \item The \textit{type} of a traffic agent: An agent can be either \textit{distracted}, thus not interacting with the ego-vehicle, or be \textit{attentive}, thus cooperatively avoid collision with the ego-vehicle.
\end{itemize}
The belief tracker is implemented as a factored histogram filter \cite{thrun2005probabilistic}. 
Each exo-agent is associated with a probability distribution over the set of possible hidden state values.
At each time step, we use \modelname to generate mean motions for an agent conditioned on all possible hidden state values. By comparing the mean motions with the actual observed motion, we compute the likelihood of the observation and update the posterior belief using the Bayes rule.


\subsection{\algname}
The core of \algname is a context-aware POMDP model solved using a state-of-the-art belief tree search algorithm (HyP-DESPOT) \cite{Hyp-despot}. We present the details of the model as follows.
\subsubsection{State and Observation Modelling}
A state in \algname includes both discrete-domain variables and continuous-domain variables:
\begin{itemize}
  \item State of the ego-vehicle, $s_c=(x,y,\phi,\mu)$, including the position $(x,y)$, heading direction $\phi$, and the intended driving path $\mu$.
  \item Observable states of exo-agents, $\{s_i=(x,y,\vec{v})\}_{i\in I_{exo}}$, including the position $(x,y)$ and the current velocity $\vec{v}$. $I_{exo}$ defines the set of indices of exo-agents. 
  \item Hidden states of exo-agents, $\{\theta_i=(t_i, \mu_i)\}_{i\in I_{exo}}$, including the type and the (sampled) intended path of the $i$th traffic agent.
\end{itemize}
We assume that the ego-vehicle can observe its own state and discretized values of the observable states of exo-agents. The hidden states of exo-agents can only be inferred and modelled with beliefs.
\subsubsection{Action Modelling}
The action space of the ego-vehicle consists of its steering angle and acceleration. Given the well-known exponential complexity of POMDP planning \cite{Kaelbling_1998}, \algname decouples the action space of the ego-vehicle to keep the branching factor of the planning problem within a tractable range. Concretely, we restrict the POMDP to compute the acceleration along the intended path, while the steering angle is generated using a pure-pursuit algorithm \cite{coulter1992purepursuit}. The action space contains three possible accelerations for each time step: $\{ACC, MAINTAIN, DEC\}$. The acceleration value for $ACC$ and $DEC$ is $ \SI{3}{\meter \per \second^2}$ and $\SI{-3}{\meter \per \second^2}$, respectively. The maximum speed of the ego-vehicle is $\SI{6}{\Ms}$.
\subsubsection{Transition Modelling \label{sec::transition}}
\algname predicts traffic agents' motion using the following set of models. Distracted traffic agents are assumed to track their intended path with the current speed. Attentive traffic agents also tend to follow the sampled path, but use \modelname to generate the actual local motion. The motion of all agents, including the ego-vehicle, are constrained by their kinematics, \eg, pedestrians are simulated using holonomic motion and car-like vehicles are simulated using bicycle models. To model stochastic transitions of exo-agents, their motion are perturbed by Gaussian noises on the displacement.
\subsubsection{Reward Modelling}
The reward function in \algname takes in to account safety, efficiency, and smoothness of driving. It assigns large penalties when the ego-vehicle collides with any exo-agent, uses a motion cost to penalize driving at low speed, and finally, penalizes frequent deceleration. Details of the reward function can be found in \cite{PORCA}.

\section{Results}
We want to answer the following questions in the experiments:
\begin{itemize}
    \item Can \simname simulate realistic dense traffic on complex maps?
    \item What are the benefits brought by \simname over rule-based models commonly used in simulators?
    \item Can \algname drive a vehicle safely and efficiently in dense unregulated urban traffic?
\end{itemize}
We provide both qualitative and quantitative results to answer these questions.


\subsection{Real-world Benchmark Scenarios}

We designed three real-world benchmark scenarios to evaluate the performance of \simname and the \algname planner. 

\begin{itemize}
    \item{Singapore-Highway (\figref{fig:benchmarks}a)}
    A highway in Singapore with multiple lanes. Traffic agents try to drive as fast as possible and thus conduct overtaking frequently. 
    \item{Magic-Roundabout (\figref{fig:benchmarks}b)}
    A roundabout at Swindon, England with very complex layout. Traffic agents meet at the main roundabout and the accompanying intersections, having to coordinate with each other.
    \item{Meskel-Intersection (\figref{fig:benchmarks}c)}
    A complex intersection at the Meskel square, Addis Abeba. Traffic agents come from different directions and encounter at the intersection, all of them driving aggressively.
\end{itemize}
Maps of the scenarios are fetched from online and imported into \simname. It then simulates unregulated traffic on these maps using \modelname.
All scenarios contain 120 heterogeneous traffic agents driving or walking in the region of interest, each conducting aggressive and unregulated behaviors. 
Once an agent moves out of the region, we replace it with new agents inside the region to maintain the high density of the traffic.
\subsection{Simulation on Benchmark Scenarios}
The realism and accuracy of our base model GAMMA have been validated in \cite{GAMMA}. In this section, we perform a qualitative study on the simulation performance of \modelname.
\figref{fig:benchmarks}(d-f) shows qualitative simulation results on the benchmark scenarios. Comparison with the real-world scenarios shows that the simulated traffic closely represent the reality. More simulation results can be found in the accompanying video or via 
\url{https://youtu.be/dNiR0z2dROg}.

\subsection{Comparison with Rule-based Simulation}
We compare \modelname with rule-based behaviours commonly used in simulators to demonstrate the sophistication of behaviours. Particularly, we compare \modelname with a reactive model that moves agents along lane center-curves and uses time-to-collision (TTC) \cite{luo2019iros} to calculate the agents' speeds. Performances of the two models are measured using the average speed of traffic agents and a congestion factor defined as the percentage of agents being jammed in the crowd, which are removed after remaining stationary for a substantial amount of time. These measures indicate how smart a behaviour model is at collision avoidance and driving efficiency, which human drivers are adept at.

\begin{figure}[!t]
\centering
\setlength{\fboxrule}{0pt}                                                              
\begin{tabular}{c}                                          
\hspace{-0.3cm} \fbox{\includegraphics[width=0.48\textwidth]{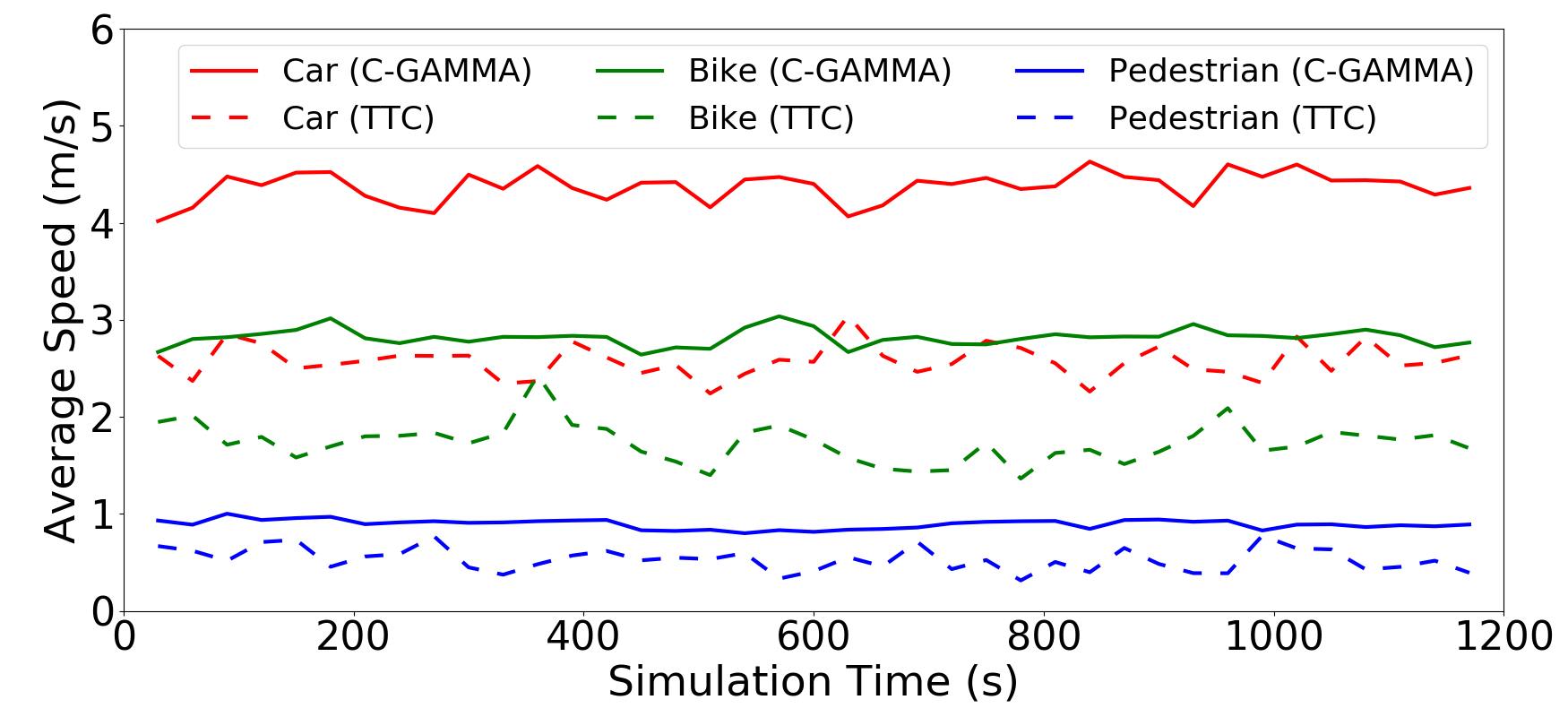}} \\
(\subfig{a}) \\
\hspace{-0.3cm} \fbox{\includegraphics[width=0.48\textwidth]{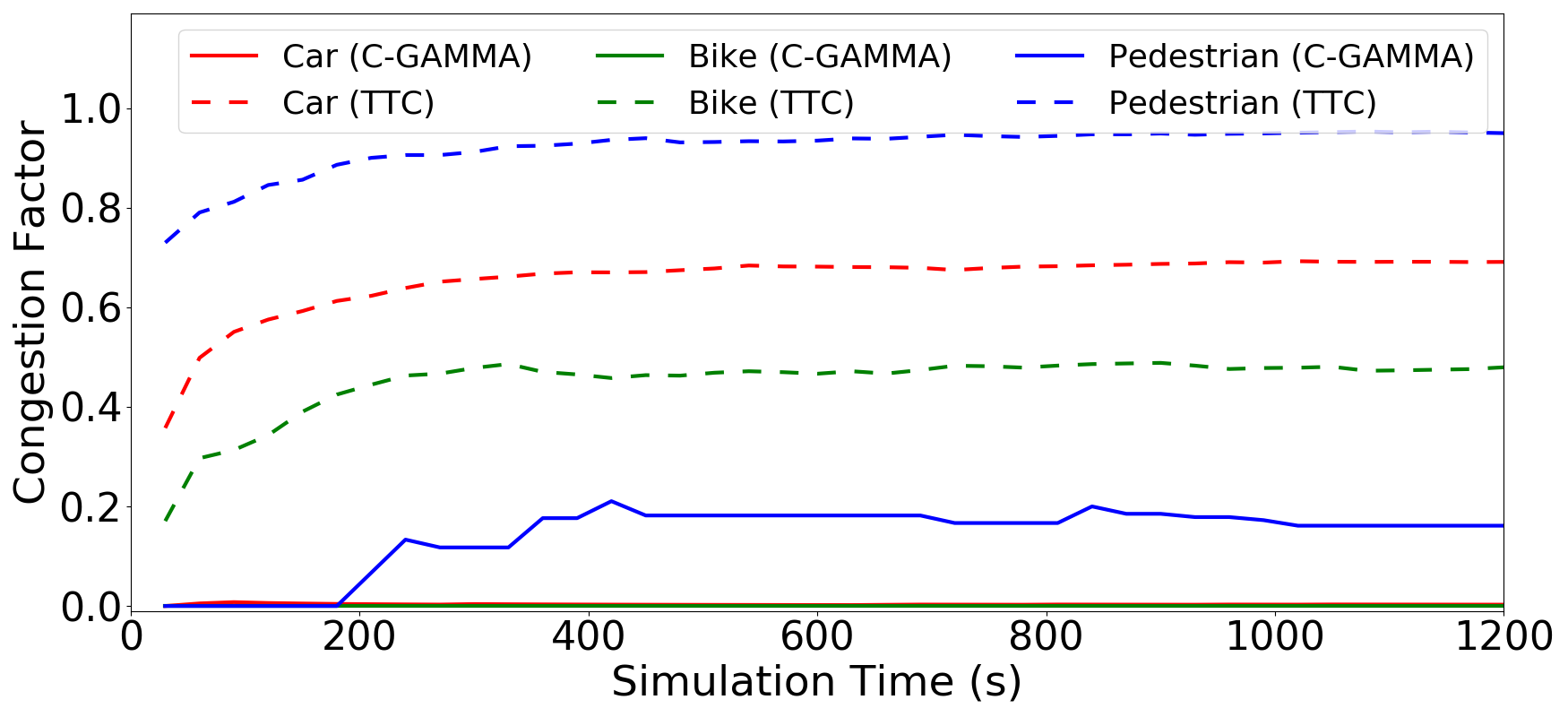}} \\
(\subfig{b})
\end{tabular}
\caption{Performance profile of \modelname and TTC on the Meskel-Intersection: (a) average speed of traffic agents; (b) congestion factor of the traffic.}   \label{fig:simprofile}     
\vspace{-0.1cm}
\end{figure}

\figref{fig:simprofile} shows a detailed profile of the agent speeds and the congestion factors for different types of agents against the simulation time. 
\modelname generates faster and smoother traffic than the TTC in all benchmark scenarios throughout 20 minutes of simulation.
The congestion factor of the TTC-controlled traffic grows quickly with the simulation time, indicating that agents fail to coordinate with each other. In contrast, \modelname consistently produces higher agent speeds and low congestion factors for all agent types. 
This is because \modelname explicitly models cooperation between agents and provides an optimal collision avoidance motion using both steering and acceleration. 


\subsection{Efficiency and Scalability of Simulation}

\begin{table}[!t]
\centering
\caption{Time performance and scalability of \modelname for simulating different number of agents on a laptop with an i7-9750H CPU and an RTX 2060 GPU.}
\begin{tabular}{ cccccccc }
\toprule
Number of Agents & 150 & 200 & 250 & 300 & 350 & 400 \\
\hline
Frequency (\si{\hertz}) & 28.4	& 20.2	& 16.6	& 12.7	& 9.4 & 7.1 \\
\bottomrule
Update Time (\si{\milli\second})  & 35.3 & 49.6 & 60.2 & 78.9 & 106.9 & 141.8 \\
\bottomrule
\end{tabular}
\label{tab::sim_frequency}
\end{table}

Efficiency tests in \tabref{tab::sim_frequency} show that \modelname scales well with the density of the crowd. The simulation runs at high rates even when modelling up to 400 agents, and the growth of computation time is almost linear until the map saturates with agents.

\subsection{Driving Performance of the \algname Planner}
We now validate the performance of the \algname planner by comparing its driving performance with local-collision avoidance and simple planning baselines.
For local collision avoidance, we directly use GAMMA to control the ego-vehicle; For simple planning, we use a roll-out algorithm that plans for optimal action by casting multiple roll-outs using a default policy. The default policy applies the following rules: accelerate the ego-vehicle when exo-agents in front are far way ($>\SI{4}{\meter}$ away), maintain half-speed when they are in caution range ($2\sim\SI{4}{\meter}$ away), and decelerate when they are close-by ($<\SI{2}{\meter}$ away). 

\tabref{tab::plan_performance} provides measurements of the collision rate per step, the average vehicle speed, and the frequency of deceleration when driving the ego-vehicle using \algname, \modelname, and Roll-out.
In summary, the simple planner drives over-conservatively, while local collision avoidance drives over-aggressively. Sophisticated planning using \algname offers a successful trade-off between aggressiveness and conservatives.

Compared to Roll-out that can barely move in the crowd, \algname can drive the vehicle through the crowd at significantly higher speed while remaining safe and smooth.
Compared to \modelname which drives aggressively but leads to high collision rate, \algname achieves similar driving speed with much safer behaviours;
We thus conclude that sophisticated long-term planning is important for crowd-driving, and \algname ensures safe, efficient, and smooth driving.

\begin{table}[!t]
\centering
\caption{Comparison on the driving performance of driving algorithms. Roll-out and \algname run at 3HZ. \modelname runs at 20 HZ. A ``step'' is counted as 1/3s.}
\begin{tabular}{ cccc }
\toprule
&\hspace{-7pt} \thead{Collision / step} &\hspace{-7pt} \thead{Avg. Speed (\si{\Ms})} &\hspace{-7pt} \thead{Dec. / step} \\
\hline
Roll-out & 0.00095 & 2.1 & 0.15 \\
\modelname & 0.002 & 5.53 & 0.12 \\
\algname & 0.00069 & 4.53 & 0.16 \\
\bottomrule
\end{tabular}
\label{tab::plan_performance}
\end{table}

\section{conclusion}
We presented \simname, a simulator for generating high-fidelity interactive data for developing, training, and testing crowd-driving algorithms. 
The simulator uses online maps to automatically construct unregulated dense traffic at any location of the world. By integrating topological road contexts with an optimization-based crowd behavior model, \simname can generate complex and realistic crowds that closely represent unregulated traffic in the real-world. We also provided Context-POMDP as a reference planning algorithm for future development. We envision that \simname will support a wide range of applications such as perception, control, planning, and learning for driving in unregulated dense urban traffic.

\section{Acknowledgement}
This research is supported in part by the Singapore MoE AcRF Tier 2 grant MOE2016-T2-2-068 and the Singapore A*STAR Undergraduate Scholarship SE/AUS/15/016.

\clearpage
\newpage
\bibliographystyle{ieeetr}      
\bibliography{main}
\end{document}